\documentclass[letterpaper]{article}
\usepackage{aaai}
\usepackage{times}
\usepackage{helvet}
\usepackage{courier}
\usepackage{graphicx}
\usepackage{amsmath}
\usepackage{amssymb}
\usepackage{booktabs}
\usepackage{algorithm}
\usepackage{algorithmic}
\usepackage{multirow}
\usepackage{makecell}
\nocopyright
\title{InsightSR: Refining Symbolic Regression Search Spaces via Parallel Semantic and Structural LLM Guidance}

\author{
Yating Ling, Wenjing Cun, Zhitang Chen \\
}

\begin{document}

\maketitle
\begin{abstract}
\begin{quote}
Symbolic regression (SR) seeks to discover parsimonious mathematical laws from observational data, yet conventional approaches often struggle with the vast combinatorial search space of physically meaningful expressions. We present InsightSR, a framework that embeds Large Language Models (LLMs) as a guiding layer around the PySR genetic programming engine. Rather than relying on LLMs to generate expressions directly, InsightSR uses LLMs to progressively transform the search space itself through two complementary pathways: a Semantic Seed Pathway that proposes dimensionally consistent functional skeletons, and a Structural Feature Pathway that recommends nonlinear feature transformations. These transformations accumulate over iterations, broadening the input space and shifting the symbolic search from constructing deep expression trees over raw variables to assembling shallow trees over a rich, semantically informed feature set. A post-generation feedback loop evaluates candidates, categorizes features by their empirical utility, and refines the guidance for the next iteration, transforming the discovery process from open-ended generation into iterative, self-correcting refinement. Across three benchmarks, InsightSR achieves a 95\% exact recovery rate on the Feynman benchmark and 80.18\% accuracy on the LLM-SRBench LSR-Transform task, substantially outperforming state-of-the-art genetic programming and neural-symbolic methods while maintaining strong out-of-distribution generalization on real-world datasets.
\end{quote}
\end{abstract}

\section{Introduction}

Symbolic regression (SR) aims to distill underlying mathematical laws directly from observed data, serving as a cornerstone for automated scientific discovery \cite{schmidt2009distilling,brunton2016discovering}. While SR produces interpretable closed-form expressions, its optimization is inherently NP-hard \cite{virgolin2022symbolic} and the combinatorial search space of candidate expressions grows exponentially with expression complexity. A further challenge is the difficulty of incorporating physical constraints, such as dimensional consistency and known conservation laws, into the search process. As a result, purely data-driven SR methods frequently yield expressions that achieve high numerical accuracy but violate basic physical principles.

Existing approaches predominantly rely on stochastic search over the discrete space of mathematical expressions. The most widely used are Genetic Programming (GP) variants \cite{koza1994genetic,cranmer2023pysr,virgolin2021improving}, which evolve populations of expression trees through mutation and crossover, and more recent methods based on Monte Carlo Tree Search (MCTS) \cite{sun2022symbolic,shojaee2023transformer}, which frame discovery as a sequential decision-making task. Although these methods can recover approximate symbolic forms, they suffer from low sample efficiency and tend to overfit under observational noise. More critically, without mechanisms to enforce physical constraints, they frequently produce expressions that are mathematically flexible but physically meaningless, lacking the parsimony and domain consistency necessary for scientific interpretation.

The integration of neural networks has further expanded the SR toolkit. Deep reinforcement learning methods such as DSR \cite{petersen2021deep} and uDSR \cite{landajuela2022unified} treat equation discovery as a policy optimization problem, using learned gradients to guide the search through expression space. Transformer-based models such as NeSymReS \cite{biggio2021neural} and E2E \cite{kamienny2022end} leverage large-scale pre-training to map data patterns directly to mathematical expressions. Other specialized approaches, such as PhySO \cite{landajuela2021discovering}, incorporate dimensional constraints into the search to enforce physical consistency. Although these neural methods accelerate discovery, they are fundamentally data-driven: their performance depends heavily on the training distribution, and they often fail to generalize to out-of-distribution regimes. Moreover, the physical constraints they enforce are typically hard-coded during training rather than dynamically reasoned about, limiting their adaptability to novel physical contexts.

Recent work has explored integrating Large Language Models (LLMs) into SR to leverage scientific priors encoded. LLM-SR~\cite{shojaee2025llmsr} generates equation skeletons via an LLM and optimizes their parameters through BFGS, storing successful patterns in an experience buffer to guide future iterations. SR-LLM~\cite{guo2025srllm} casts SR as a reinforcement learning problem with an LLM-based policy network, coupled with retrieval-augmented generation for incremental knowledge accumulation. LaSR~\cite{li2024lasr} focuses on LLM-guided feature engineering, recommending transformations to expand the input space. PiSR via LLM~\cite{taskin2026pisr} incorporates LLM-derived evaluations into the loss function as a physics-informed regularization term. While promising, these methods are often constrained by two limitations. First, they incur substantial inference overhead, as they require frequent LLM queries throughout the search process. Second, and more importantly, each method employs LLM guidance along a single dimension---either structural hypothesis generation or input space transformation---without exploiting the potential synergy of integrating both. This separation overlooks a key opportunity: structural priors can inform which features to engineer, and engineered features can simplify the structural forms needed to fit the data.

To address these limitations, we propose InsightSR, a framework that embeds LLMs as a guiding layer around the PySR genetic programming engine~\cite{cranmer2023pysr}. Rather than relying on the LLM to generate candidate expressions directly, InsightSR uses the LLM to progressively transform the search space through two complementary pathways. A Semantic Seed Pathway proposes dimensionally consistent functional skeletons that provide a physics-informed warm-start, while a Structural Feature Pathway recommends nonlinear feature transformations that accumulate over generations, progressively broadening the input space. This shifts the burden of the search from constructing deep expression trees over raw variables to assembling shallow combinations over a semantically enriched feature set. A learning-from-results feedback loop closes each generation: the LLM evaluates the Pareto-optimal candidates, categorizes features by their empirical utility, and updates a knowledge base that informs both pathways in the subsequent iteration, transforming symbolic discovery from open-ended generation into iterative, self-correcting refinement.

Our work makes three key contributions:

\begin{enumerate}
\item \textbf{Semantic Seed Pathway.} We introduce a mechanism that leverages domain-aware LLMs to propose dimensionally consistent functional skeletons that serve as physics-informed seeds for the evolutionary search. By auditing and reconfiguring candidate topologies to align with target physical units, this pathway prunes the search space of physically inconsistent expressions before numerical fitting begins. The LLM also provides per-operator complexity biases that steer the genetic programming engine toward physically plausible operator combinations.

\item \textbf{Structural Feature Pathway.} We design a complementary pathway in which the LLM recommends nonlinear feature transformations based on the problem context and the performance history of prior generations. These transformations accumulate across iterations, progressively broadening the input space. This shifts the evolutionary search from constructing deep expression trees over raw variables to assembling shallow combinations over an enriched feature basis, substantially reducing the structural complexity that the symbolic engine must resolve.

\item \textbf{Closed-Loop Iterative Refinement.} We implement a learning-from-results mechanism that closes the discovery loop. After each generation, the LLM evaluates candidates along multiple dimensions including numerical accuracy, physical interpretability, and feature utility. These assessments are accumulated in a dynamic knowledge base that informs both the semantic and structural pathways in subsequent iterations, enabling the search to self-correct and progressively converge toward the optimal symbolic form.
\end{enumerate}

\section{Method}

\begin{figure*}[t]
\centering
\includegraphics[width=0.99\textwidth]{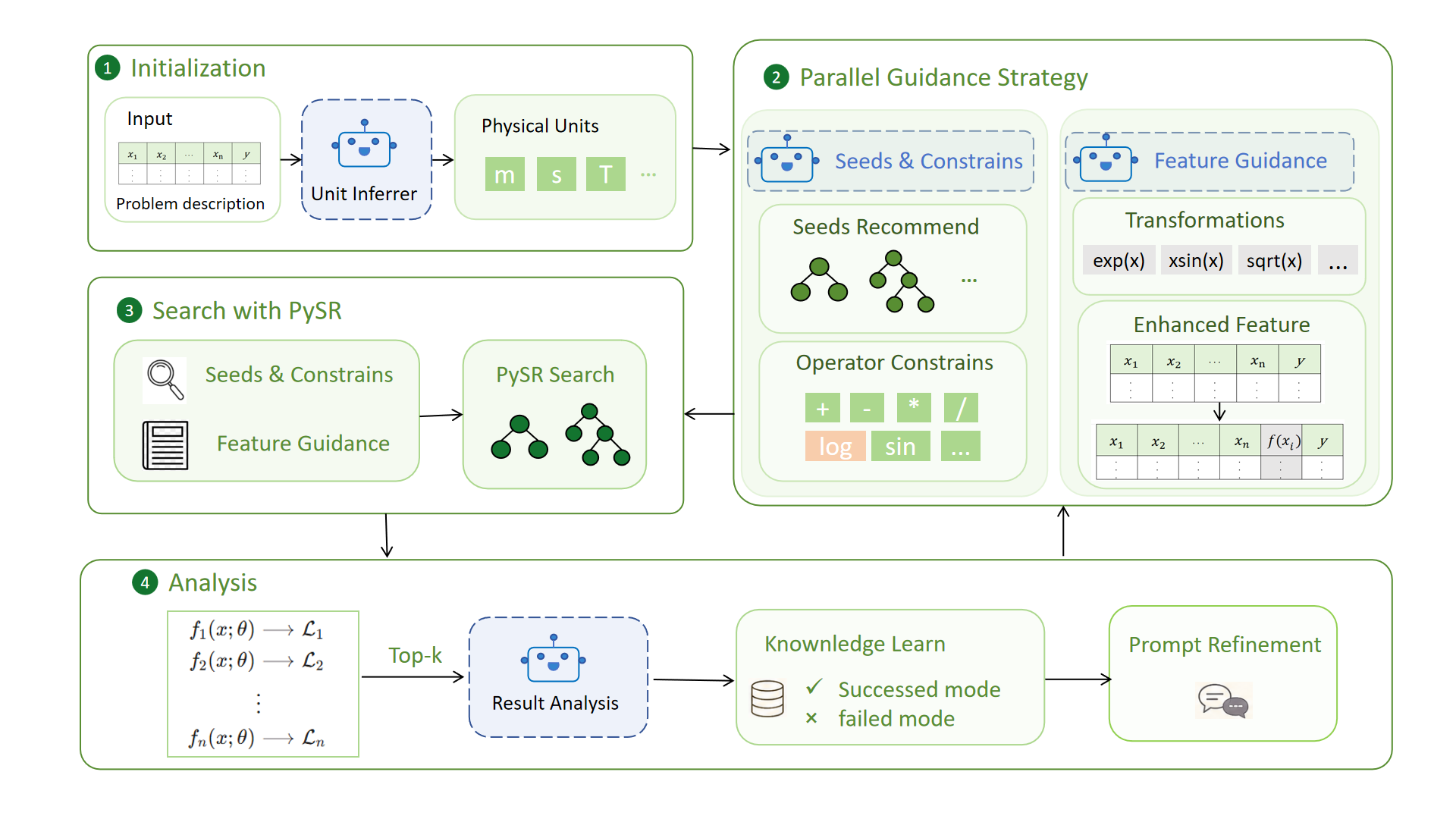}
\caption{The system architecture of InsightSR. The framework follows a four-phase iterative loop per generation: (1) \textbf{Contextual Initialization}, where the LLM performs automated unit synthesis and domain analysis; (2) \textbf{Parallel Guidance Strategy}, which concurrently generates seed expressions (Strategy A) and recommends feature enhancements (Strategy B); (3) \textbf{Search with PySR}, where the integrated heuristics steer the PySR engine; and (4) \textbf{Strategic Analysis}, where the LLM evaluates the top-$K$ candidates and formulates a plan for the next iteration.}
\label{fig:architecture}
\end{figure*}

The goal of SR is to discover an analytical expression $f: \mathbb{R}^d \to \mathbb{R}$ that accurately describes the relationship in a dataset $\mathcal{D} = \{(\mathbf{x}_i, y_i)\}_{i=1}^N$. Following the standard GP formulation, we treat this as a multi-objective optimization that balances numerical accuracy against structural parsimony:

\begin{equation}
\min_{f \in \mathcal{F}} \mathcal{L}(f, \mathcal{D}) = \text{MSE}(y, f(\mathbf{X})) + \lambda \cdot \mathcal{C}(f),
\end{equation}

where $\mathcal{C}(f)$ denotes the structural complexity of $f$ (measured by the number of nodes in its expression tree) and $\lambda$ controls the parsimony penalty.

InsightSR organizes the discovery process into a four-phase iterative loop, summarized in Figure~\ref{fig:architecture} and formalized in Algorithm~\ref{alg:llm4sr}. Each generation embeds LLM-derived domain knowledge at three intervention points---before, during, and after the evolutionary search---shifting PySR from undirected stochastic exploration toward purposefully guided optimization. We detail each phase below.

\begin{algorithm}[t]
\caption{InsightSR Search Procedure}
\label{alg:llm4sr}
\small
\begin{algorithmic}
\REQUIRE Dataset $(\mathbf{X}, y)$, problem context $\mathcal{C}$, units $\mathcal{U}$
\ENSURE Optimal symbolic expression $f^*$
\STATE Initialize knowledge base $\mathcal{K} \gets \emptyset$
\FOR{$g = 1$ to $G$}
  \STATE // Dual-Path Heuristic Guidance
  \STATE $\mathcal{F}_{seed} \gets \text{\sc LLM.SemanticSeeds}(\mathcal{C}, \mathcal{U}, \mathcal{K})$ \hfill $\triangleright$ Seed Pathway
  \STATE $\mathbf{X}_{aug} \gets \text{\sc LLM.StructuralFeatures}(\mathcal{C}, \mathcal{K})$ \hfill $\triangleright$ Feature Pathway
  \STATE $\mathcal{G} \gets \text{MergeGuidance}(\mathcal{F}_{seed}, \mathbf{X}_{aug})$
  \STATE $\text{candidates} \gets \text{\sc PySR.GuidedSearch}(\mathbf{X}_{aug}, y, \mathcal{G})$
  \STATE $\text{eval}, \text{insights} \gets \text{\sc LLM.StrategicAnalysis}(\text{candidates})$
  \STATE $\mathcal{K} \gets \text{\sc KB.Accumulate}(\text{insights}, \text{eval})$
  \IF{$\text{Loss}(f^*) < 10^{-10}$}
    \STATE \textbf{break}
  \ENDIF
\ENDFOR
\RETURN Best expression $f^*$ from candidates
\end{algorithmic}
\end{algorithm}

\subsection{Contextual Initialization and Unit Synthesis}

The framework first resolves the physical dimensions of each variable. It extracts available dimensional information from the dataset metadata and problem description. For variables whose units remain ambiguous or unspecified, an LLM infers probable dimensions based on the semantic context of the problem and the physical role of the target variable. The resulting unit assignments persist as dimensional constraints throughout the discovery loop. By enforcing dimensional homogeneity, the framework prunes physically inconsistent candidates from the search space, ensuring that the symbolic search remains grounded in the underlying physics.

\subsection{Parallel Guidance Strategy}
The core of InsightSR is a parallel guidance strategy in which the LLM simultaneously contributes along two complementary dimensions: top-down semantic seeding and bottom-up feature engineering.

\textbf{Semantic Seed Pathway.} This pathway leverages domain-aware LLMs to propose dimensionally consistent functional skeletons that serve as physics-informed seeds for the evolutionary search. Given the problem metadata and the unit constraints resolved in Section~3.1, the LLM generates candidate expression topologies whose dimensional consistency is audited before they enter the population. Each skeleton contains symbolic constants that are subsequently optimized numerically. By seeding the population with physically plausible structures, this pathway prunes a vast region of dimensionally inconsistent expressions from the search space before the evolutionary optimization begins.

\textbf{Structural Feature Pathway.} In parallel, this pathway enriches the input space through LLM-guided feature engineering. At each generation, the LLM analyzes the performance history and identifies high-utility nonlinear transformations $\phi(\mathbf{x})$ based on patterns observed in surviving candidates. These transformations are appended to the input matrix:
\begin{equation}
\mathbf{X}_{aug} = [\mathbf{x}_1, \dots, \mathbf{x}_d, \phi_1(\mathbf{x}), \dots, \phi_k(\mathbf{x})],
\end{equation}
where $\phi_j(\mathbf{x})$ includes operations such as power-law terms $x_i^n$ and transcendental mappings $\sin(x_i)$. Crucially, these features accumulate across generations, progressively broadening the input space. This shifts the search from constructing deep expression trees to discover complex nonlinear relationships to assembling shallow combinations over an already-enriched feature set, substantially reducing the structural depth that the evolutionary engine must resolve.

The two pathways converge within the PySR engine. Rather than searching the unconstrained space $\mathcal{F}$ over raw inputs $\mathbf{x}$, the optimizer operates over an informed manifold:
\begin{equation}
f^* = \arg\min_{f \in \mathcal{F}_{seed}} \mathcal{L}(f(\mathbf{X}_{aug}), \mathcal{D}).
\end{equation}
Semantic seeds provide macro-level structural constraints grounded in physics, while engineered features supply micro-level building blocks derived from empirical patterns. Together, they confine the search to a physically plausible region of the expression space without sacrificing the numerical flexibility needed for high-precision fitting.

\subsection{Search with PySR}

The synthesized guidance consists of three components: symbolic seeds, operator preferences, and the augmented feature matrix $\mathbf{X}_{aug}$. These are passed to a modified PySR engine for evolutionary search. To steer the search toward LLM-informed functional structures, we introduce a complexity-biasing mechanism that adjusts the structural penalty of each expression tree. The adjusted complexity $\mathcal{C}'(f)$ is computed as the sum of baseline operator costs $c_o$ modified by an LLM-provided bias $\omega_o$:
\begin{equation}
\mathcal{C}'(f) = \sum_{o \in f} (c_o + \omega_o).
\end{equation}
Operators recommended by the LLM receive a negative bias $\omega_o$, granting them a complexity discount that promotes their survival on the Pareto front. Conversely, operators deemed physically implausible are penalized with a positive bias, increasing their effective cost and suppressing their propagation across generations. In addition, we employ a warm-start mechanism that retains the best candidate from the previous generation as an elite member of the new population. This ensures that prior refinements are preserved and built upon, enabling progressive convergence toward the optimal symbolic form.

\subsection{Multi-Dimensional Evaluation and Strategic Feedback}
After each evolutionary search completes, the LLM serves as a post-generation critic to evaluate the Pareto-optimal candidates and guide the next iteration. The evaluation synthesizes multiple criteria: numerical accuracy, physical interpretability, and the completeness with which the expression incorporates the relevant variables. In parallel, the system performs a diagnostic analysis of the feature space, categorizing each engineered transformation as effective or ineffective based on its contribution to reducing the residual error. These assessments are stored in a dynamic knowledge base, enabling the framework to learn from both successful patterns and failure modes across generations. The knowledge base then informs the next iteration in three ways: it refines the seed expressions proposed by the Semantic Seed Pathway, adjusts the operator complexity biases used by the PySR engine, and prioritizes feature transformations for the Structural Feature Pathway. By closing the loop between empirical search and semantic reasoning, the framework progressively narrows the search toward physically meaningful expressions until convergence.

\section{Experimental Setup}

\subsection{Benchmarks}
We evaluate on three benchmarks summarized in Table~\ref{tab:benchmarks}.

\begin{table*}[]
\caption{Benchmark datasets overview}
\label{tab:benchmarks}
\centering
\small
\begin{tabular}{lccc}
\toprule
Benchmark & Problems & Example\\
\midrule
Feynman \cite{udrescu2020ai} & 100 & $\frac{m_0}{\sqrt{1-v^2/c^2}}$ \\
LLM-SRBench \cite{shojaee2025llm} &240 & $-kA(t)^2 + \frac{k_z A(t)^2}{\beta A(t)^4 + 1}$ \\
Real-World \cite{shojaee2025llmsr} & 4  &0.8 $\sin(x) - 0.5v^3 - 0.2x^3 - 0.5xv - x \cos(x)$\\
\bottomrule
\end{tabular}
\end{table*}

The \textbf{Feynman 100} benchmark~\cite{udrescu2020ai} consists of 100 physics equations derived from the Feynman Lectures on Physics, covering classical mechanics, electromagnetism, quantum mechanics, and thermodynamics. Each problem has a known ground-truth expression, enabling exact recovery evaluation.

The \textbf{LLM-SRBench}~\cite{shojaee2025llm} comprises five categories: LSR-Transform and LSR-Synth across Physics, Chemistry, Biology, and Material Science. Each category features different data-generating mechanisms and varying complexity, assessing cross-domain generalization.

The \textbf{Real-World} benchmark~\cite{shojaee2025llmsr} includes four real-world datasets (Oscillator 1, Oscillator 2, E. coli, Stress-Strain). Each dataset provides both in-distribution (ID) and out-of-distribution (OOD) test splits, enabling evaluation of out-of-distribution generalization.

\subsection{Implementation Details}
Table~\ref{tab:hyperparams} lists the key hyperparameters used in our experiments.

\begin{table}[t]
\caption{Hyperparameters}
\label{tab:hyperparams}
\centering
\small
\begin{tabular}{lc}
\toprule
Parameter & Value \\
\midrule
LLM & Qwen-3.5 27B\\
LLM Temperature & 0.3 \\
PySR Iterations & 100--500 \\
LLM Generations & 3--30 \\
Populations & 30 \\
Population Size & 50 \\
Parsimony & 0.003 \\
Max Expression Size & 20--25 \\
Early Stop Threshold & $10^{-10}$ \\
Loss Function & MSE \\
\bottomrule
\end{tabular}
\end{table}

\subsection{Evaluation Metrics}
The Coefficient of Determination ($R^2$) measures the proportion of variance captured by the model, defined as:
\begin{equation}
R^2 = 1 - \frac{\sum_{i=1}^{n} (y_i - \hat{y}_i)^2}{\sum_{i=1}^{n} (y_i - \bar{y})^2}
\end{equation}
where $y_i$ represents the ground truth, $\hat{y}_i$ is the predicted value, and $\bar{y}$ is the empirical mean. Consistent with our optimization objective, the Normalized Mean Squared Error (NMSE) is defined as $\text{NMSE} = 1 - R^2$.

The Accuracy under Tolerance ($\text{acc}_{\tau}$) is defined as a binary success metric for each problem. A discovery is considered successful ($1$) only if the maximum relative error across all $n$ test samples is within the threshold $\tau$; otherwise, it is marked as failure ($0$):
\begin{equation}
\text{acc}_{\tau} = \mathbb{I} \left( \max_{1 \le i \le n} \left| \frac{y_i - \hat{y}_i}{y_i} \right| \le \tau \right)
\end{equation}
where $\mathbb{I}(\cdot)$ is the indicator function.
The Exact Recovery Rate is defined as a special case where the symbolic form is mathematically equivalent to the ground truth, typically resulting in $\text{acc}_{\tau} \to 1$ for $\tau \to 0$.

\section{Results}

\subsection{Feynman Benchmark}
\begin{figure}[t]
\centering
\includegraphics[width=0.85\linewidth]{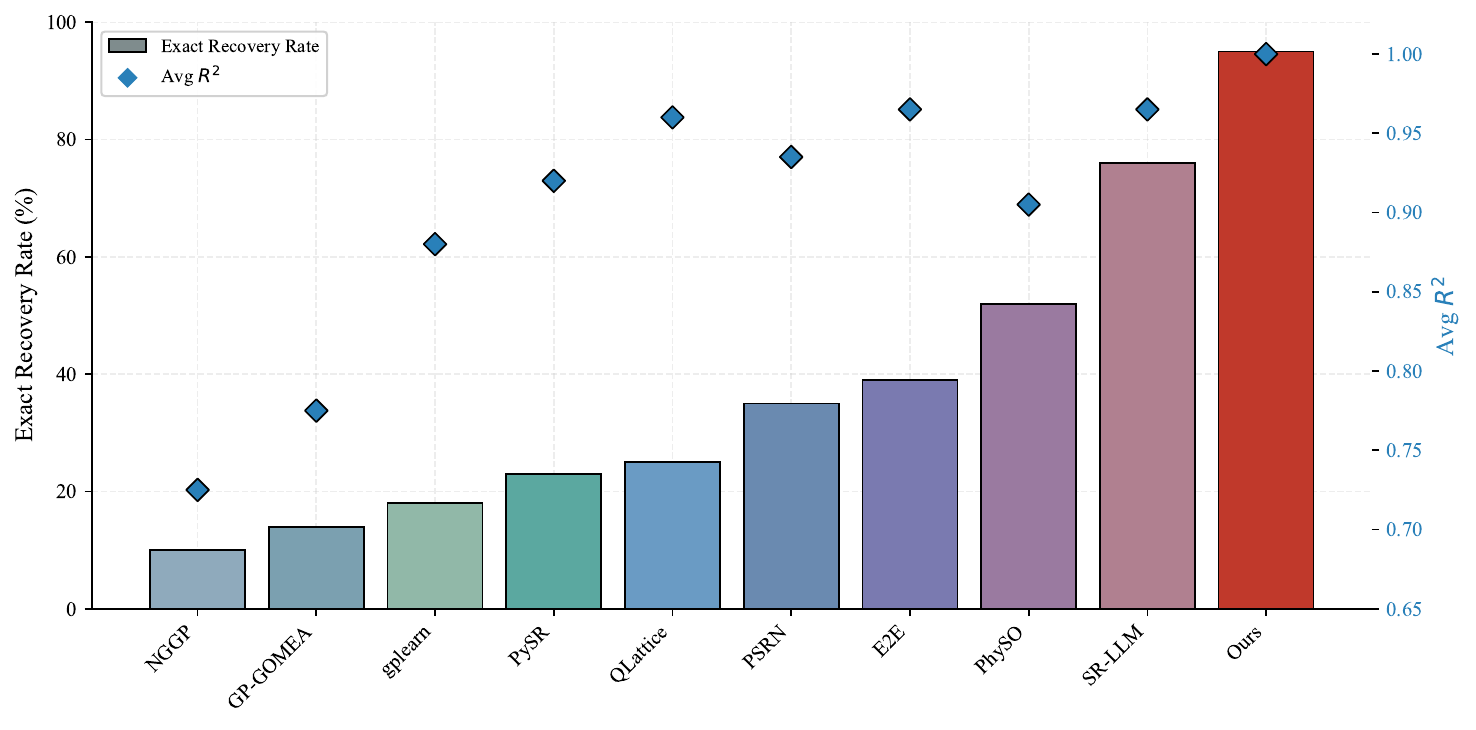}
\caption{Feynman benchmark: exact recovery rate (bars) with average R$^2$ overlay (diamonds).}
\label{fig:feynman}
\end{figure}

\begin{figure}[t]
\centering
\includegraphics[width=0.8\linewidth]{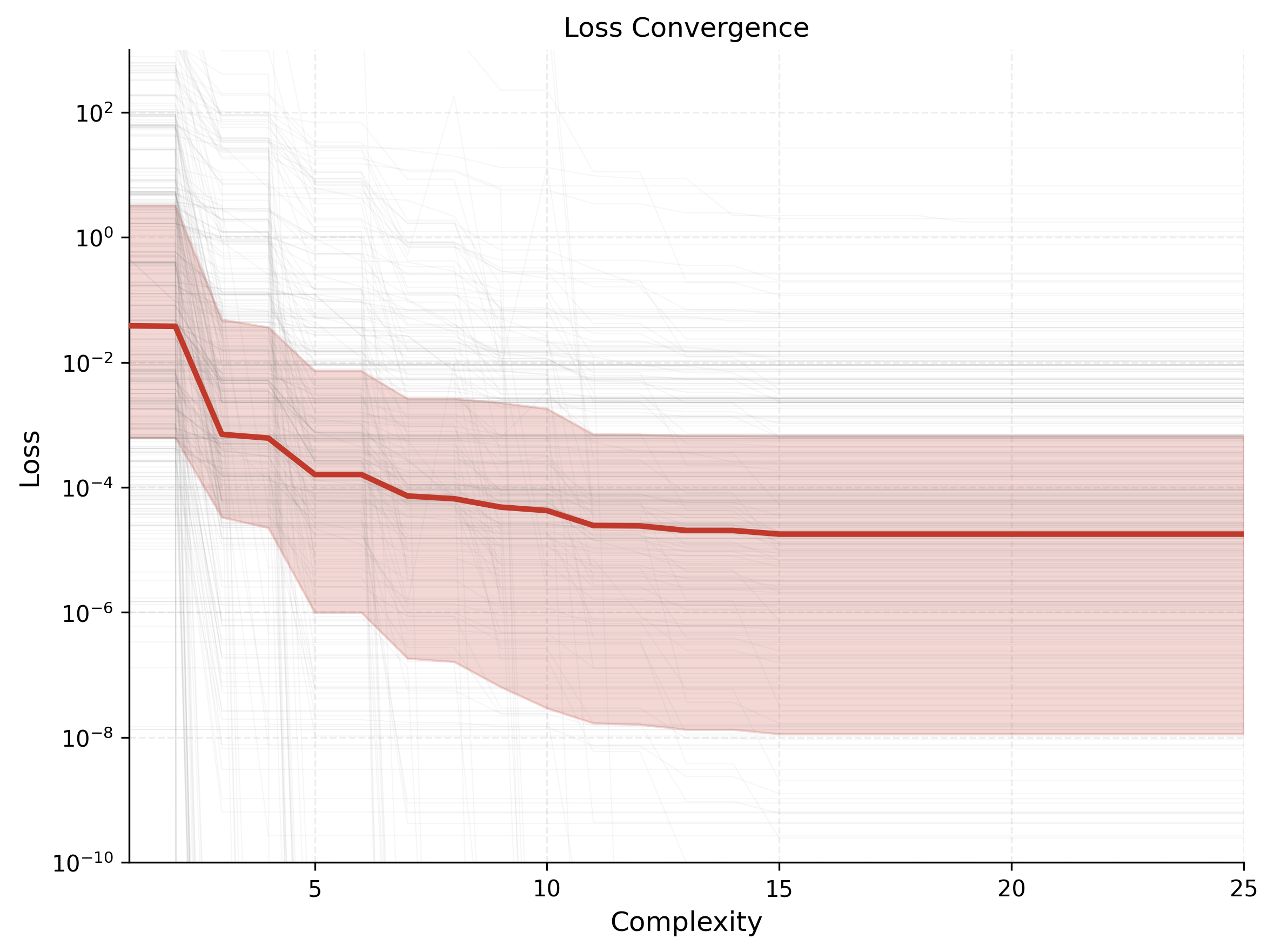}
\caption{Loss vs. expression complexity Pareto front. The solid line shows the median; shaded region shows the 25th--75th percentile.}
\label{fig:convergence}
\end{figure}

Figure~\ref{fig:feynman} summarizes the performance on the Feynman benchmark. InsightSR achieves a 95\% exact recovery rate and an average $R^2$ of 0.9999, outperforming the previous state of the art. Figure~\ref{fig:convergence} shows the Pareto front of loss versus expression complexity. The median loss declines consistently as complexity increases, and the narrowing interquartile range at higher complexity levels indicates enhanced numerical stability. We further evaluated robustness to data perturbations. As shown in Table~\ref{tab:noise_acc}, accuracy remains high under moderate noise, with Acc$_{0.1}$ reaching 79\% at a noise level of 0.001.

\begin{table}[h]
\centering
\caption{Noise vs Accuracy Performance}
\label{tab:noise_acc}
\begin{tabular}{cc}
\toprule
\textbf{noise} & \textbf{Acc$_{0.1}$ (\%)} \\
\midrule
0.05  & 37 \\
0.01  & 68 \\
0.005 & 72 \\
0.001 & 79 \\
0     & 97 \\
\bottomrule
\end{tabular}
\end{table}

\subsection{LLM-SRBench}

We further evaluate our method on the LLM-SRBench across five scientific domains. As summarized in Table~\ref{tab:srbench}, our method achieves a substantial gain in the LSR-Transform category, reaching an Acc$_{0.1}$ of 80.18\% compared to the previous best of 50.45\% (LaSR).

\begin{figure}[t]
\centering
\includegraphics[width=0.85\linewidth]{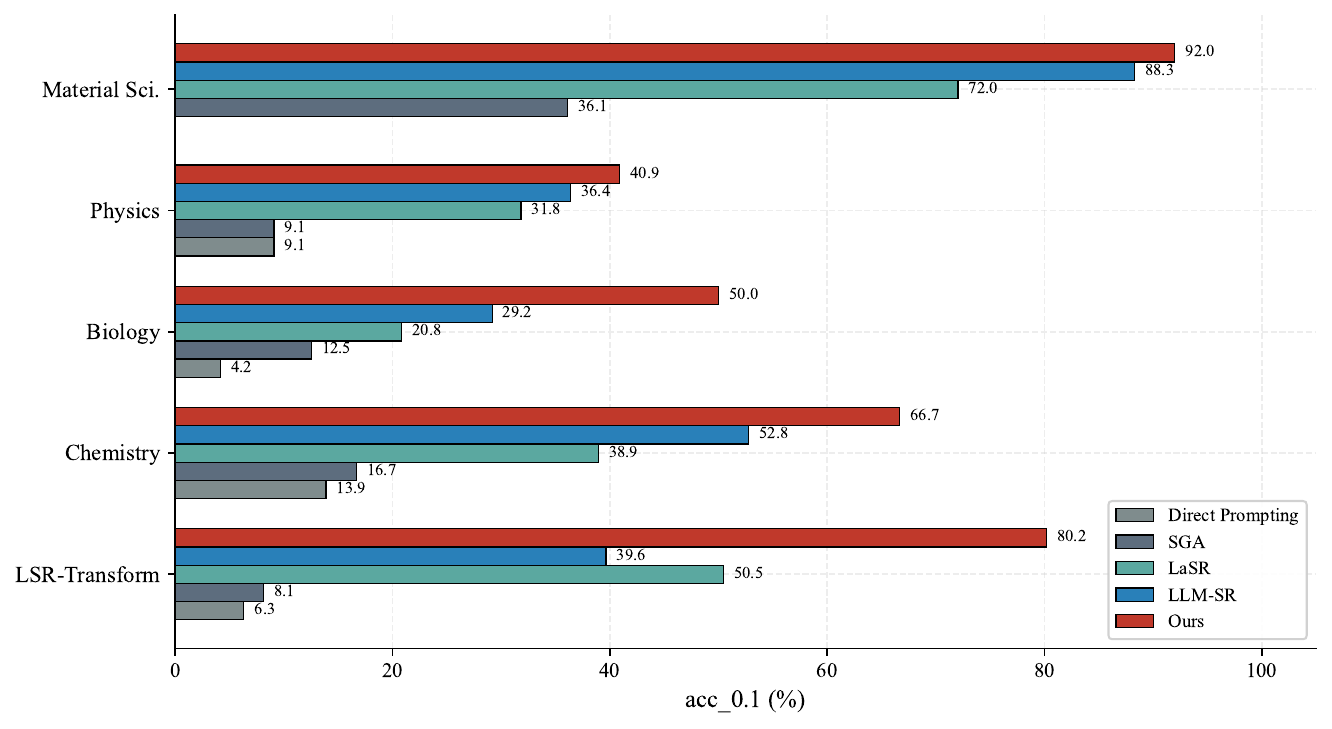}
\caption{LLM-SRBench Acc$_{0.1}$ comparison across five categories.}
\label{fig:srbench}
\end{figure}

Performance is consistent across the four domain-specific synthetic datasets. Specifically, our method achieves an accuracy of 66.67\% in Chemistry, 50.00\% in Biology, 40.91\% in Physics, and 92.00\% in Material Science. In terms of numerical precision, our method yields NMSE values that are several orders of magnitude lower than other LLM-based solvers. These results are further illustrated in Figure~\ref{fig:srbench}, which shows the relative performance across all methods.

\begin{table*}[t]
\caption{LLM-SRBench results. }
\label{tab:srbench}
\centering
\setlength{\tabcolsep}{2.5pt}
\footnotesize
\begin{tabular}{lcccccccccc}
\toprule
\multirow{2}{*}{Method} & \multicolumn{2}{c}{LSR-Transform} & \multicolumn{2}{c}{Chemistry} & \multicolumn{2}{c}{Biology} & \multicolumn{2}{c}{Physics} & \multicolumn{2}{c}{Material Sci.} \\
\cmidrule(lr){2-3} \cmidrule(lr){4-5} \cmidrule(lr){6-7} \cmidrule(lr){8-9} \cmidrule(lr){10-11}
 & Acc$_{0.1}$ $\uparrow$ & NMSE$\downarrow$ & Acc$_{0.1}$$\uparrow$ & NMSE$\downarrow$ & Acc$_{0.1}$$\uparrow$ & NMSE$\downarrow$ & Acc$_{0.1}$$\uparrow$ & NMSE$\downarrow$ & Acc$_{0.1}$$\uparrow$ & NMSE$\downarrow$ \\
\midrule
Direct Prompting & 6.31 & 0.263 & 13.88 & 2.2e-2 & 4.16 & 0.465 & 9.09 & 0.065 & 0.00 & 0.048 \\
SGA              & 8.11 & 0.232 & 16.66 & 5.5e-4 & 12.51 & 0.013 & 9.09 & 0.051 & 36.11 & 6.0e-4 \\
LaSR             & 50.45 & 0.001 & 38.92 & 9.1e-5 & 20.83 & 1.5e-4 & 31.81 & 9.9e-4 & 72.04 & 9.2e-6 \\
LLM-SR           & 39.64 & 0.009 & 52.77 & 4.1e-6 & 29.16 & 3.1e-6 & 36.36 & 7.6e-5 & 88.28 & 3.2e-9 \\
\midrule
\textbf{Ours}    & \textbf{80.18} & \textbf{1.3e-14} & \textbf{66.67} & \textbf{8.0e-7} & \textbf{50.00} & \textbf{7.3e-7} & \textbf{40.91} & \textbf{1.6e-5} & \textbf{92.00} & \textbf{4.7e-8} \\
\bottomrule
\end{tabular}
\end{table*}

\subsection{Real-World Datasets}

Table~\ref{tab:llmsr} compares the NMSE results of our method against several baselines on four real-world datasets, covering both in-distribution (ID) and out-of-distribution (OOD) scenarios. Overall, our method obtains the lowest NMSE in most test cases.

\begin{figure}[t]
\centering
\includegraphics[width=0.85\linewidth]{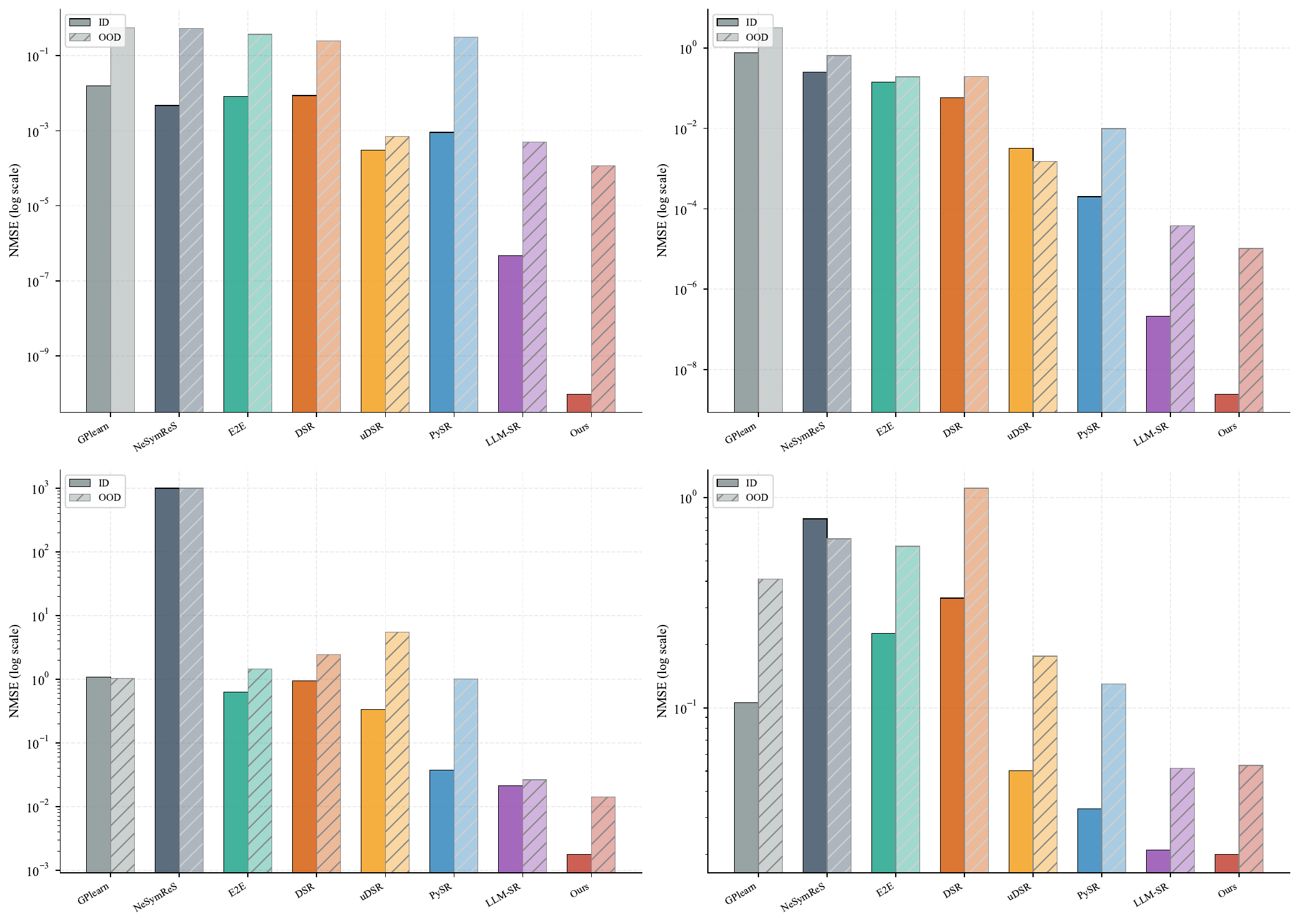}
\caption{LLM-SR benchmark NMSE comparison.}
\label{fig:llmsr}
\end{figure}

On the Oscillator 1 and Oscillator 2 datasets, our method achieves ID NMSE values of $9.55 \times 10^{-11}$ and $2.45 \times 10^{-9}$, respectively, representing an improvement of several orders of magnitude over the previous best LLM-based method (LLM-SR). For the E. coli dataset, our method yields an ID NMSE of $1.76 \times 10^{-3}$ and an OOD NMSE of $1.41 \times 10^{-2}$. In the Stress-Strain task, our results are $2.00 \times 10^{-2}$ (ID) and $5.32 \times 10^{-2}$ (OOD), which are comparable to or better than LLM-SR. These results are further illustrated in Figure~\ref{fig:llmsr}.

\begin{table*}[t]
\caption{Real-world datasets results. ID: in-distribution, OOD: out-of-distribution.}
\label{tab:llmsr}
\centering
\small
\begin{tabular}{lcccccccc}
\toprule
\multirow{2}{*}{Method} & \multicolumn{2}{c}{Oscillator 1} & \multicolumn{2}{c}{Oscillator 2} & \multicolumn{2}{c}{E. coli} & \multicolumn{2}{c}{Stress-Strain} \\
\cmidrule(lr){2-3} \cmidrule(lr){4-5} \cmidrule(lr){6-7} \cmidrule(lr){8-9}
 & ID & OOD & ID & OOD & ID & OOD & ID & OOD \\
\midrule
GPlearn & 1.55e-02 & 5.57e-01 & 7.55e-01 & 3.19e+00 & 1.08e+00 & 1.04e+00 & 1.06e-01 & 4.09e-01 \\
NeSymReS & 4.70e-03 & 5.38e-01 & 2.49e-01 & 6.47e-01 & --- & 1.45e+00 & 7.93e-01 & 6.38e-01 \\
E2E & 8.20e-03 & 3.72e-01 & 1.40e-01 & 1.91e-01 & 6.32e-01 & 1.45e+00 & 2.26e-01 & 5.87e-01 \\
DSR & 8.70e-03 & 2.45e-01 & 5.80e-02 & 1.95e-01 & 9.45e-01 & 2.43e+00 & 3.33e-01 & 1.11e+00 \\
uDSR & 3.00e-04 & 7.00e-04 & 3.20e-03 & 1.50e-03 & 3.32e-01 & 5.46e+00 & 5.02e-02 & 1.76e-01 \\
PySR & 9.00e-04 & 3.11e-01 & 2.00e-04 & 9.80e-03 & 3.76e-02 & 1.01e+00 & 3.31e-02 & 1.30e-01 \\
LLM-SR & 4.65e-07 & 5.00e-04 & 2.12e-07 & 3.81e-05 & 2.14e-02 & 2.64e-02 & 2.10e-02 & 5.16e-02 \\
\midrule
\textbf{Ours} & \textbf{9.55e-11} & \textbf{1.15e-04} & \textbf{2.45e-09} & \textbf{1.03e-05} & \textbf{1.76e-03} & \textbf{1.41e-02} & \textbf{2.00e-02} & \textbf{5.32e-02} \\
\bottomrule
\end{tabular}
\end{table*}

\subsection{Qualitative Case Study}
To provide an intuitive understanding of our method's performance, we compare the ground-truth equations against the discovered expressions across all three benchmarks. Table~\ref{tab:qualitative} highlights how our method recovers the core functional forms, with matching terms bolded in the discovered column.

\begin{table*}[t]
\caption{Ground truth vs. discovered equations. Matching terms are bolded.}
\label{tab:qualitative}
\centering
\small
\begin{tabular}{clll}
\toprule
Benchmark & Problem & Ground Truth & Discovered \\
\midrule
\multirow{2}{*}{Feynman}
 & I.6.2a & $\displaystyle\frac{e^{-\theta^2/2}}{\sqrt{2\pi}}$ & $0.3989 \cdot \mathbf{e^{-\theta^2/2}}$ \\
 & I.8.14 & $\sqrt{(x_2-x_1)^2+(y_2-y_1)^2}$ & $\mathbf{\sqrt{(x_1-x_2)^2+(y_1-y_2)^2}}$ \\
\midrule
\multirow{2}{*}{LLM-SRBench}
 & II.6.15b & $\displaystyle\frac{8\pi E_f\epsilon r^3}{3\sin(2\theta)}$ & $\displaystyle\frac{4\pi}{3} \cdot \frac{\mathbf{E_f\epsilon r^3}}{\mathbf{\sin\theta\cos\theta}}$ \\
 & I.48.2 & $-c\sqrt{1-c^4m^2/E_n^2}$ & $-\mathbf{c}\sqrt{1-\mathbf{m^2c^4/E_n^2}}$ \\
\midrule
\multirow{2}{*}{Real-World}
 & Oscillator 1 & \makecell[t]{$0.8\sin x - 0.5v^3$ \\ $- 0.2x^3 - 0.5xv - x\cos x$} & \makecell[t]{$-0.5(\mathbf{v^3}+\mathbf{xv})$ \\ $+ \mathbf{\sin x}(-0.275\mathbf{\cos x}+0.075)$} \\
 & Oscillator 2 & $\makecell[t]{0.3\sin t - 0.5v^3 \\ - xv - 5xe^{0.5x}}$ & \makecell[t]{$0.3\mathbf{\sin t} - 0.5\mathbf{v^3}$ \\ $- \mathbf{xv} + \mathbf{e^x}(x^6-5) + \text{const}$} \\
\bottomrule
\end{tabular}
\end{table*}

\section{Discussion}

The experimental results across three benchmarks demonstrate that InsightSR addresses two central challenges in symbolic regression: the combinatorial explosion of the search space and the difficulty of maintaining physical consistency under purely data-driven optimization. The dual-pathway design contributes along both dimensions. The Semantic Seed Pathway provides a physics-informed warm-start, which is reflected in the 95\% exact recovery rate on the Feynman benchmark: by seeding the population with dimensionally consistent skeletons, the search bypasses vast regions of physically meaningless expressions that would otherwise dominate the early generations of a standard GP run.

The Structural Feature Pathway addresses the depth bottleneck inherent in GP-based SR. By accumulating LLM-recommended transformations across generations, InsightSR progressively broadens the input space, shifting the search from constructing deep expression trees to assembling shallow combinations over an enriched feature basis. This reduction in required structural depth is directly reflected in the 80.18\% accuracy on the LSR-Transform task, where target expressions exhibit complex functional forms that would otherwise require prohibitively deep trees to discover. The two pathways operate synergistically: semantic constraints narrow the space of candidate structures, while engineered features simplify the structural complexity needed to fit the data, together enabling the PySR engine to operate far more effectively than in its standalone configuration.

The performance on real-world datasets, such as the Oscillator and E. coli benchmarks, further highlights the framework's capacity for out-of-distribution generalization. Unlike purely neural symbolic methods that may overfit to specific training distributions, the learning-from-results feedback mechanism in InsightSR helps the evolved expressions remain parsimonious and physically interpretable. The complexity-biasing mechanism, informed by the LLM's strategic analysis, prioritizes operators that align with established domain knowledge. This adaptive regularization is crucial for scientific discovery, where the ultimate objective is to identify governing principles that remain valid beyond the observed data range.

While the integration of LLM guidance enhances search efficiency, the framework's reliance on initial metadata suggests a potential sensitivity to problem descriptions. In scenarios where variable semantics are ambiguous, the accuracy of the unit synthesis module becomes a critical factor. The current ``learning-from-results'' loop remains computationally viable by strategically querying the LLM between generations rather than at every search step. Future research may investigate the use of multi-modal priors to further refine the initialization phase and enhance the framework's robustness in the absence of explicit textual metadata.

\section{Conclusion}
This paper presented InsightSR, a framework that embeds Large Language Models as a guiding layer around the PySR genetic programming engine. Through two complementary pathways, semantic seeding and structural feature engineering, the framework progressively transforms the search space, shifting symbolic regression from deep tree construction over raw variables to shallow assembly over an enriched feature basis. A closed-loop feedback mechanism enables the system to learn from search outcomes and refine its guidance across generations. Extensive evaluations demonstrate that InsightSR achieves state-of-the-art accuracy across three benchmarks while maintaining strong out-of-distribution generalization on real-world datasets. These results suggest that strategic integration of LLM reasoning with established evolutionary search engines offers a practical and effective path toward automated scientific discovery.

\bibliographystyle{aaai}
\bibliography{references}

\end{document}